\documentclass{article}
\usepackage{amsmath}
\usepackage{graphicx}
\usepackage{algorithm2e}
\usepackage{enumitem}
\usepackage{subfig}
\usepackage[utf8]{inputenc} 
\usepackage[T1]{fontenc}    
\usepackage{hyperref}       
\usepackage{url}            
\usepackage{booktabs}       
\usepackage{amsfonts}       
\usepackage{nicefrac}       
\usepackage{microtype}      
\usepackage{xcolor}         
\usepackage{wrapfig}        
\usepackage{adjustbox}      
\usepackage{amssymb}
\usepackage{float}

\DeclareMathOperator*{\argmax}{argmax}

\title{Synthesizing Diverse Network Flow Datasets with Scalable Dynamic Multigraph Generation}

\author{
    Arya Grayeli, Vipin Swarup\footnote{Corresponding author: swarup@acm.org}, Steven E. Noel\\
    The MITRE Corporation, McLean, Virginia, USA \\
}

\begin{document}

\maketitle

\begin{NoHyper}%
\renewcommand{\thefootnote}{~}
\footnote{Approved for Public Release; Distribution Unlimited. Public Release Case Number 25-0980.}
\footnote{©2025 The MITRE Corporation. ALL RIGHTS RESERVED.}
\renewcommand{\thefootnote}{footnote}
\end{NoHyper}%

\begin{abstract}
Obtaining real-world network datasets is often challenging because of privacy, security, and computational constraints. In the absence of such datasets, graph generative models become essential tools for creating synthetic datasets. In this paper, we introduce a novel machine learning model for generating high-fidelity synthetic network flow datasets that are representative of real-world networks. Our approach involves the generation of dynamic multigraphs using a stochastic Kronecker graph generator for structure generation and a tabular generative adversarial network for feature generation. We further employ an XGBoost (eXtreme Gradient Boosting) model for graph alignment, ensuring accurate overlay of features onto the generated graph structure. We evaluate our model using new metrics that assess both the accuracy and diversity of the synthetic graphs. Our results demonstrate improvements in accuracy over previous large-scale graph generation methods while maintaining similar efficiency. We also explore the trade-off between accuracy and diversity in synthetic graph dataset creation, a topic not extensively covered in related works. Our contributions include the synthesis and evaluation of large real-world netflow datasets and the definition of new metrics for evaluating synthetic graph generative models.
\end{abstract}

\section{Introduction}
\vspace{-1.5mm}

Graphs are meaningful abstractions for representing complex relationships among entities in a domain. Correspondingly, graph datasets capture the variety in these relationships, locally as well as globally. The recent growth of development in graph algorithms has driven the need for high-fidelity graph datasets that closely mimic real-world phenomena. However, observing real-world phenomena is often time-consuming and difficult. The ability to generate synthetic graph datasets alleviates the limitations of relying on solely empirical data, thus motivating work in graph dataset generation.

In the cyber domain, network flows (or netflows) represent the timing and type of information exchanged between machines. It is natural to structure network flow information as a large graph. However, obtaining real-world netflow datasets for research purposes is challenging because of privacy, security, and computational concerns~\cite{lanl}. The release of such data can compromise the reference networks and make them vulnerable to cyberattacks. Simply obfuscating IP addresses or adding noise to the network is insufficient to address these privacy concerns. Moreover, recording all netflows requires significant infrastructure and the data cannot be easily transferred from one network to another. Therefore, there is a need for graph generative models that can create synthetic datasets of netflows for research purposes.

To address these challenges, we introduce a machine learning model to generate diverse ensembles of graphs (containing IP addresses and netflows) that resemble a reference graph. Our model aims to produce dynamic multigraphs that include the categorical edge attribute of port-protocol. To achieve scalability and efficiency, the model breaks down the modeling process into three stages (inspired by~\cite{nvidia}): (1)~structure generation, (2)~feature generation, and (3)~graph alignment.

Structure generation reduces the reference dynamic multigraph into a static simple graph using a stochastic Kronecker graph generator. Feature generation employs a Conditional Tabular Generative Adversarial Network (CTGAN) model~\cite{ctgan} to learn the distribution of the edge features. Graph alignment utilizes an XGBoost model~\cite{Chen2016XGBoost} to overlay the features onto the graph to determine the source and destination of each edge. Because of the separation into three stages, our model can generate graphs with varying numbers of nodes and edges. Tuning CTGAN allows for adjusting network dynamics.

This paper defines new metrics to evaluate the quality of synthetic graphs and provides experimental results to demonstrate the effectiveness of the proposed metrics and model. We evaluate our model on real-world data, thus demonstrating the value of our approach for cybersecurity research. We also improve and expand on existing works in large-scale graph dataset generation.

We summarize our main contributions as follows:
\begin{enumerate}[itemsep=0.1ex] 
    \item Synthesis and evaluation on large real-world netflow datasets, providing a scalable, accurate, and generalizable netflow dataset synthesizer.
    \item New characteristic-free metrics for both structural and dynamic properties to evaluate accuracy and diversity of dataset graph generation models.
    \item Efficient extension of previous approaches to model multigraph dynamics.
    \item Improvement on structural results from previous work, yielding more complex models while maintaining efficiency.
\end{enumerate}

\vspace{-1.5mm}
\section{Related Work}
\vspace{-1.5mm}

Since network flows can be represented as graphs, we approach the problem from a graph generation perspective. Graph generative models started as purely mathematical models that generated random graphs~\cite{random}, scale-free graphs~\cite{scale_free}, and small world graphs~\cite{small_world}. As the field progressed, more complex mathematical models such as Recursive Matrix (R-MAT)~\cite{rmat} were developed.

In recent years, machine learning approaches to graph generation have begun. Most commonly used models involve Graph Neural Networks (GNNs)~\cite{gnn3,gnn,gnn2}. Many employ autoregressive models that generate on a node, edge, or block basis. However, these models have difficulty scaling to larger graphs with thousands of nodes. Thus, there lies a gap between efficient but simple algorithms that make many assumptions and more generalizable but slower deep learning approaches.

Furthermore, most such models lack the ability to generate edge and node features, because the problem of joint structure and feature generation is complex. Also, previous work has generally employed static simple graphs, even though most real-world networks have a dynamic multigraph structure~\cite{dynamic}. While the problems are fundamentally the same, some different considerations need to be made.

There is a lack of standardized metrics for graph generation that are free from biased assumptions, such as fitting to degree distribution. Additionally, existing measures often fail to consider dynamics and feature distributions. In the context of synthetic dataset generation, there is even less emphasis on evaluating the diversity and information gain provided by each new dataset. Instead, most evaluations focus primarily on the model's accuracy with respect to the reference data.

In the area of network flow generation, there have been attempts to synthetically generate network traffic data, but primarily using tabular formats instead of a graph structure. Most commonly used are variational autoencoders~\cite{vae} and generative adversarial networks (\cite{gan_netflow2},\cite{gan_netflow}).

There have also been recent works in diffusion models~\cite{diffusion}. While these models may be a more scalable deep learning approach to generating graphical data, they tend to lose information regarding the underlying structure of the graphs which results in highly concentrated networks that are significantly different from the actual structure needed. This is undesirable since most users of this data care about the connectivity of the network, e.g., for tasks such as modeling adversarial attack paths or link prediction.

\vspace{-1.5mm}
\section{Methods}
\vspace{-0.5mm}

This section describes our approach to synthetic netflow graph generation. Subsection~\ref{subsec:problem_definition} (Problem Definition) sets the goal of generating diverse and realistic graph datasets that replicate a reference graph, while emphasizing scalability and diversity. Subsection~\ref{subsec:structure_generation} (Structure Generation) describes the application of a stochastic Kronecker graph generator to efficiently model a network's underlying topology. Subsection~\ref{subsec:feature_generation} (Feature Generation) outlines the application of the CTGAN model to generate edge features that reflect the reference dataset's distributions. Subsection~\ref{subsec:graph_alignment} (Graph Alignment) details the use of an XGBoost model to match generated features with structural edges, ensuring a coherent representation of netflows.

\subsection{Problem Definition}
\label{subsec:problem_definition}
\vspace{-0.5mm}

The goal of this model is to sample a diverse set of graphs similar to a reference graph $\mathcal{G} = (\mathcal{V}, \mathcal{E})$. Formally, $\mathcal{V}$ is the set of $N$ IP addresses (nodes) and $\mathcal{E}$ is the set of $M$ netflows (edges). For network flow generation, there can exist multiple edges between pairs of nodes, with each edge containing a start and end time and a port-protocol combination. This representation of netflows aligns with the widely used flow 5-tuple in previous work~\cite{lanl}. Thus, the probabilistic model must generate dynamic multigraphs with the additional categorical edge feature of port-protocol.

Because of the large number of edges in netflow graphs, we emphasize scalable and efficient models. We also uphold (1) similarity as we want datasets that model real-world phenomena and (2) diversity because we want to add new data to this domain. For example, defining a model that returns the same reference graph would have perfect similarity but no diversity, thus rendering it useless in applications of the tool. We define metrics to evaluate the quality of synthetic graph datasets in Section~\ref{section:metrics}.

\begin{figure}[H]
    \centering
    \includegraphics[width=1.0\textwidth]{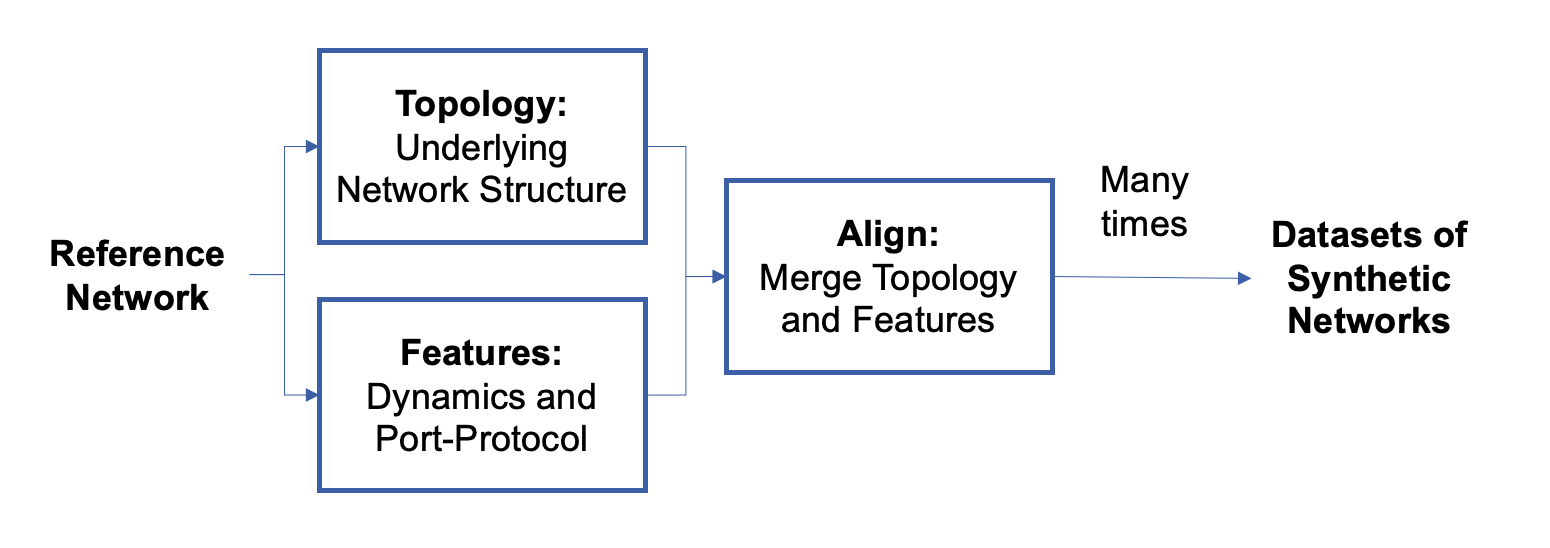}
    \caption{\label{fig:approach_overview} Overview of the model approach for generating synthetic netflow graphs. The process is divided into three stages: structure generation using a stochastic Kronecker graph, feature generation with CTGAN, and graph alignment via XGBoost. This modular framework enables scalable and efficient synthesis of diverse and realistic network datasets.}
\end{figure}

With the framework from~\cite{nvidia} as inspiration, we likewise break the modeling process into three steps: structure generation, feature generation, and graph alignment. This is illustrated in Figure~\ref{fig:approach_overview}.

By making structure and feature generation independent processes, the problem becomes simpler, which allows for more efficient solutions. This aligns with our focus on applications to large scale, real-world datasets.

Once the topology and dynamics are computed separately, we align their outputs to unify into the resulting generated graph. Each of these steps rely on a random process. Thus, by executing this process many times, we can generate an ensemble of datasets that each represent unique and realistic netflows.

\vspace{-0.5mm}
\subsection{Structure Generation}
\label{subsec:structure_generation}
\vspace{-0.5mm}

Before learning the graph topology, we reduce the reference dynamic multigraph into a static simple graph. By doing so, we convert the problem of modeling $M$ dynamic edges into modeling an $N \times N$ binary adjacency matrix. This saves compute as netflow graphs generally have many flows between the same two nodes ($M \gg N^2$). Thus, this step of structure generation is not to compute which edges will be present and when. Rather, it is to lay a foundation as to which nodes are able to communicate with one another, while the later stages determine the timing of when they do so (perhaps never).

To determine this underlying topology, we apply a stochastic Kronecker graph generator~\cite{kron}. This process generates an $N \times N$ probability matrix $\hat{A} =A^{\otimes k} $, where $A$ is an $N_1 \times N_1$ probability matrix, $k = \lfloor \log_{N_1}N \rfloor$, and $\otimes$ is the tensor product of two matrices (also called the Kronecker product). Subsequently, each entry $p_{ij} \in \hat{A}$ is the probability of an edge existing between nodes $i$ and $j$. One could calculate $\hat{A}$ from $A$ and then sample edges using the probabilities in all $N^2$ entries in the adjacency matrix to extract the final binary adjacency matrix in $O(N^2)$ time.

We can speed up the Kronecker graph sampling process to approximately $O(E)$ time with only a small loss in accuracy, where $E$ is the number of edges in the reference network if it were a static simple graph. This is because of the recursive nature of the Kronecker product, since each $p_{ij} \in \hat{A}$ is the product of $k$ decisions to choose a $p_{ij} \in A$. This allows us to simulate the process to generate a binary adjacency matrix with $E$ edges, as shown in Algorithm~\ref{alg:fast_kron}.

\begin{algorithm}
    \SetAlgoLined
    \KwIn{$A, N_1, k, E$}
    \KwOut{$\hat{A}$}

    $A \gets A / \sum_{ij} A_{ij}$\;
    $e \gets 0$\;
    \While{$e < E$} {
        $i,j \gets 0,0$\;
        \For{$t \gets 1$ \KwTo $k$} {
            $i_t,j_t \gets $ randomly choose entry in $A$ using $p = A_{i_tj_t}$\;
            $i \gets i + i_t * N_1^t$\;
            $j \gets j + j_t * N_1^t$\;
        }

        \If{$\hat{A}_{ij} == 0$} {
            $\hat{A}_{ij} \gets 1$\;
            $e \gets e + 1$\;
        }
    }
    \caption{Fast Kronecker graph generation algorithm. This algorithm efficiently generates a binary adjacency matrix for a graph by recursively applying the Kronecker product, reducing computational complexity to approximately $O(E N_1^2 \log N)$. It balances speed and accuracy, making it suitable for large-scale network synthesis.}
    \label{alg:fast_kron}
\end{algorithm}

This gives a sampling method in $O(E N_1^2 \log N)$ time compared to $O(N^2)$, where $N^2 \gg E$ because of the sparsity of netflows when multi-edges are removed. Note that this is an extension of the R-MAT approach in~\cite{nvidia}, which is just the special case of a $2 \times 2$ Kronecker matrix ($N_1 = 2$). Our approach allows for any size Kronecker matrix to be tested, in which we optimize over $N_1$ to find the best size. We use Kronecker graphs to model topology over other approaches as it is computationally more efficient which works better for graphs where $N$ is large. Because of the modular framework of this model, it could be substituted out for other algorithms if desired.

This process does allow for duplicate edges, i.e., multi edges. In such cases, we just rerun the process to generate another edge. Previous experiments with this approach found real-world networks to be sparse enough such that collisions only happen about 1\% of the time~\cite{kron}.

This generative process has two parameters to be optimized during training: $A$ and $N_1$. For a reference simple graph $G$, we use the KronFit algorithm~\cite{kron} to find the optimal $A$ of a given size $N_1$. Rather than optimizing on a specific metric in most approaches, KronFit directly aims to match the reference adjacency matrix using maximum likelihood estimation. This means the synthetic data is node-aligned with the reference data, which is helpful for downstream tasks and evaluation.

 Formally, KronFit solves the following:

    \[ \argmax_{A}{P(G|A)} \]

This is optimized upon via gradient descent over the log-likelihood loss $l(A) = \log P(G|A)$. The assumption is that standard stochastic gradient descent are applicable to this problem, assuming that there are relatively few local maxima. Because of the difficulty of optimizing over graphs because of their permutation invariance, the original KronFit algorithm efficiently approximates the log-likelihood and its gradient with permutation sampling in $O(E)$ time. A worry with this approach is the possibility of overfitting, as we are fitting directly to a single graph. To provide some regularization, we can choose $N_1$ appropriately as it has a direct relationship with the complexity of the learned $A$, and subsequently its chance of overfitting. The authors of KronFit deemed the Bayes Information Criterion (BIC) as appropriate for choosing $N_1$:

    \[ BIC(N_1) = -l(A_{N_1}) + \frac{1}{2} N_1^2 \log(N^2) \]

By choosing the $N_1$ with the minimum BIC, we can provide some assurance that $A$ is both complex enough and not overfit.

\vspace{-0.5mm}
\subsection{Feature Generation}
\label{subsec:feature_generation}
\vspace{-0.5mm}

While the structure of the underlying topology is being generated, the distributions of the edge features (start time, end time, and port-protocol) are also calculated. To remove some of the constraints of end times occurring after the start time, the end time calculation is simplified to the duration of the flow.

Ideally, the feature model should learn not to output a negative duration, but some post-processing can clean it up. Because of the tabular nature of the features, we employ the CTGAN model. For the encoding of both continuous and discrete features, we apply the mode-specific normalization proposed in CTGAN to the continuous values (start time and duration) and one-hot encode the discrete values (port-protocol). From there, CTGAN is trained to learn the feature distributions, from which we sample $M$ times to get the features of our synthetic flows for inference.

\vspace{-0.5mm}
\subsection{Graph Alignment}
\label{subsec:graph_alignment}
\vspace{-0.5mm}

Once the underlying structure is found and $M$ edge features are proposed, we must overlay the features onto the graph to determine the source and destination of each edge feature. For doing so, we take the encoded edge features from the CTGAN model along with a set of structural features of each node and edge from the Kronecker graph, and feed into a XGBoost model. These structural features include centrality measures of the source and destination nodes such as degree, betweenness, eigenvector, and laplacian centrality, as well as edge-based measures such as edge betweenness.
During inference, each proposed edge feature $f_j$ will be matched to an edge from the Kronecker graph $e_i$ by using an XGBoost model which models $P(e_i | f_j) \propto P(e_i \land f_j)$. Since we are doing this to choose an $e_i$ for each $f_j$, the $P(f_j)$ term becomes irrelevant so all we need to do is model $P(e_i \land f_j)$ for all $e_i$ and then take a weighted random sample to find the $e_i$ to match with $f_j$.

However, learning a model for $P(e_i \land f_j)$, for both $e$ and $f$ that are different from the training distribution, is difficult. Thus, we simulate:
    \[ P(e_i \land f_j) \approx \frac{ \sum_{f \in k_i} S(f_j, f)}{ |k_i| } = \frac{f_j \cdot (\sum_{f \in k_i} \frac{f}{|f|})}{|f_j|*|k_i|} \]
where $S$ is the cosine similarity function and $k_i$ is the set of all features for each flow in $e_i$ in the reference graph. We do this for all possible pairs of $e_i$ and $f_j$ to create a labeled dataset of probabilities for both seen and unseen flows. However, this takes $O(M^2)$ time. To reduce the complexity, notice that the sum of cosine similarities can be partially pre-computed, giving the expression on the right which can be computed in $O(MN^2)$ time. A standard XGBoost model is used to train on and predict $P(e_i \land f_j)$ given a vectorized $f_j$ and a vectorized encoding of the structural features of $e_i$. Using $P(e_i \land f_j)$, we can model and sample from $P(e_i | f_j)$ to choose a structural edge for each feature netflow.

Because of the large number of flows that need to be fit, we also employ a threshold in which values less than the threshold become a probability of zero. This is because the small variability arising from the uncertainty of the model can lead to unrealistic choices being made. Since this is a slower process, for larger datasets we can sample only a certain percentage of the possible $N^2$ structural edges to make its constant factor more optimal. With a sufficiently large $N$, we find that it does not take away from the integrity of the aligner.

\vspace{-1.5mm}
\section{Metrics and Evaluation}
\vspace{-1.5mm}
\label{section:metrics}

As mentioned in the problem definition, we emphasize three overall properties: accuracy, diversity, and scalability. The evaluation of scalability has already been discussed in the complexity analysis in the Methods sections, in which we have defined a model that is approximately linear in the total number of netflows $M$. As this is the most significant term when scaling to larger netflow datasets, we determine no need for further evaluation of scalability.

Let $G^{ref}$ be a given reference graph. Let $S$ be the resulting synthetic data output after running our model many times. Because of the periodicity of network flows, we define $T$ on a daily scale, and define each $G_t \in \mathbb{R}^{N \times N}$ as the aggregated number of active flows between each two nodes for day $t$. Then $G \in S$ is the ordered set of all $G_t$ for $t \in T$ and is the result of an independent synthetic generation by the model.

We can define addition of $G, G' \in S$ as $(G + G')_t = G_t + G'_t$ for all $t \in T$. Subsequently, $\Bar{S}$, which is the sample mean of $S$, is the following:
\begin{equation*}
   \Bar{S} = \frac{1}{|S|} \sum_{G \in S} G
\end{equation*}

For accuracy and diversity, we define the edit-distance sum $\mathcal{L}$:
\begin{equation*}
    \mathcal{L}(G, G') = \sum_{t \in T} ||G_t - G'_t||
\end{equation*}
In particular, $\mathcal{L}$ is the sum of the edit distances between the aggregated states of the graph at various points in time. We can use the edit distance directly because of the node correspondence properties of our generative model.

By defining $\mathcal{L}$ as the distance between two graphs, we can define a metric space of solutions. We seek these solutions to be both accurate and diverse, so in this space we visualize it as a high-dimensional sphere of solutions. The accuracy of the sphere is calculated in $\mathcal{A}$:
\begin{equation*}
    \mathcal{A}(G^{ref}, S) = \mathcal{L}(G^{ref}, \Bar{S})
\end{equation*}
Here the synthetic networks, or the points, are first averaged to get the sample mean. This is then compared with the true mean, i.e., the reference network. Since we want the sphere to be centered at the reference network, the distance between the sample mean and true mean is our measure of accuracy with a minimal distance being more accurate.

On the other hand, we also desire diversity in the dataset. This is reflected in $\mathcal{D}$, in which we take the standard deviation of the distance of each point in our synthetic dataset from the reference network:
\begin{equation*}
    \mathcal{D}(G^{ref}, S) = Std\{\mathcal{L}(G, G^{ref}) : \forall G \in S\}
\end{equation*}
By minimizing this value, we can ensure that each entry is closer to the surface of a sphere of solutions rather than the interior. Solutions that are more concentrated on the surface are also further from the center (i.e., the reference network) which implies greater diversity.

To measure the amount by which these points are different from the reference network, we also need a measure of the radius of such a sphere. For that, we define $\mathcal{R}$:
\begin{equation*}
    \mathcal{R}(G^{ref}, S) = \sqrt{\frac{\sum_{G \in S} \mathcal{L}(G, G^{ref})^2}{|S| - 1}}
\end{equation*}
By maximizing $\mathcal{R}$, we get additional insurance that the model does not converge on the identity function, i.e., returning the reference network. Thus, we introduce this new evaluation schema for jointly optimizing dataset accuracy and diversity.

The configuration for the evaluation of our model is inspired by high-dimensional Gaussian spheres. In this context, with high probability, points get concentrated on the surface but still retain their center at the mean. This is a reflection of maximal diversity with high accuracy, which would ensure meaningful, realistic, and diverse results. We find that similar measures of accuracy and diversity (such as using $\Bar{S}$ instead of $G^{ref}$ in the evaluation of $\mathcal{D}$ and $\mathcal{R}$) give conclusions almost identical to our metrics, regardless of model or reference network.

Thus, we introduce a new evaluation scheme for dataset accuracy and diversity metrics, inspired by high-dimensional Gaussian spheres. We incorporate both dynamics and structure without any underlying assumptions except the time blocks in $T$. Future work is needed to efficiently incorporate the non-dynamic edge features into these metrics, as we currently disregard the port-protocol.

Combining these metrics into a singular value, we seek to minimize $\mathcal{A}$ and $\mathcal{D}$ and maximize $\mathcal{R}$. However, both $\mathcal{A}$ and $\mathcal{D}$ are dependent on $\mathcal{R}$, as it constitutes the scale of values. Thus, we scale both of them by $\mathcal{R}$. We then define bias and variability as the following:
\[Bias=\frac{\mathcal{A}(G^{ref}, S)}{\mathcal{R}(G^{ref}, S)}\]
\[Variability=\frac{\mathcal{D}(G^{ref}, S)}{\mathcal{R}(G^{ref}, S)}\]
Here, the term ``variability'' does not denote variability in the traditional sense. Instead, it is a construct specific to our framework, intended to capture a relationship between the spread of the generated dataset around the reference network (\(\mathcal{R}\)) and the spread around the dataset's own mean (\(\mathcal{D}\)). This ratio is used for training our model so that generated datasets have more diversity with respect to the reference network.

In defining overall generalization error, we draw inspiration from the bias-variability trade-off, and the roles of bias and variability in the error. We thus have the generalization error \(\mathcal{E}\):
\[\mathcal{E}(G^{ref}, S) = Bias^2 + Variability = \left[\frac{\mathcal{A}(G^{ref}, S)}{\mathcal{R}(G^{ref}, S)} \right]^2 + \frac{\mathcal{D}(G^{ref}, S)}{\mathcal{R}(G^{ref}, S)}\]
Here, bias represents the systematic error in the synthetic dataset's mean relative to the reference graph. Squaring bias emphasizes the importance of minimizing this error. Then variability represents the spread of synthetic graphs around their sample mean. Including variability linearly reflects its role as a measure of diversity, which is encouraged but not prioritized over accuracy.

\begin{figure}[h]
    \centering
    \includegraphics[width=0.73\textwidth]{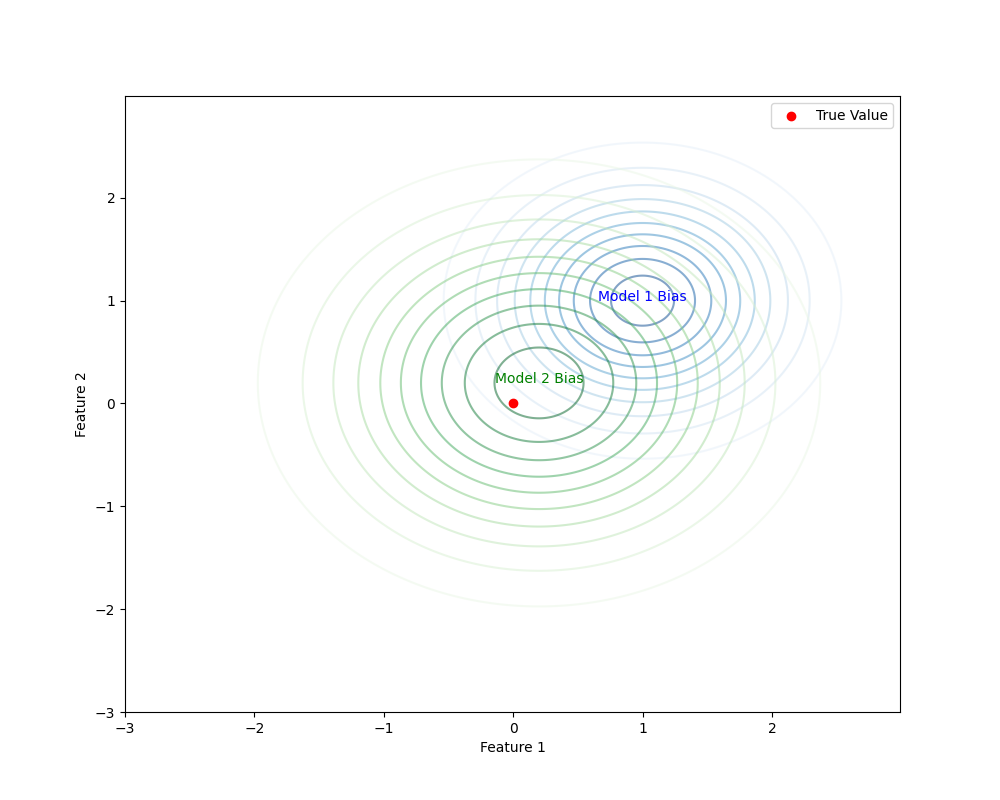}
    \caption{Conceptual illustration of the accuracy-diversity trade-off using two-dimensional Gaussian distributions. The figure contrasts two models: Model 1 (blue contours) with higher bias and lower diversity (higher variability), and Model 2 (green contours) with lower bias and higher diversity (lower variability). The red point represents the true value, highlighting the balance between achieving accuracy and maintaining diversity in model predictions.}
    \label{fig:illustrate_error_in_two_d}
\end{figure}

Figure~\ref{fig:illustrate_error_in_two_d} is a visual representation of the bias-variability trade-off. This figure serves as a simplified analogy to our concept of high-dimensional Gaussian spheres. It highlights the difference between achieving accuracy (low bias) and maintaining diversity (low variability) in model predictions. The contour lines indicate the probability density of predictions, with higher density regions suggesting more likely predictions by each model.

Figure~\ref{fig:illustrate_error_in_two_d} uses two-dimensional Gaussian distributions to illustrate the predictions of two hypothetical models. In this figure, the red point at the origin (labeled 'True Value') represents the target point that both models aim to predict (i.e., fidelity to the reference network). Model~1 (blue contours) has higher bias (poorer accuracy) and lower diversity, which corresponds to a higher variability as defined in our framework. Its mean is more offset from the true value, and its predictions are tightly clustered. Conversely, Model~2 (green contours) has lower bias (better accuracy) and higher diversity, corresponding to a lower variability. Its mean is closer to the true value, and it has a wider spread of predictions.

Beyond the primary evaluation metrics, we report performance at intermediate stages before the structure and feature are overlaid. In particular, for evaluating the underlying generated graph structure, we use the degree similarity, average shortest path length, effective diameter, number of distinct edges, density, and the clustering coefficient as secondary measures. For evaluating the sampled features, we employ Cumulative Distribution Function (CDF) plots. These align with more standard measures to show there is still consistency between those and our proposed metrics.

\vspace{-1.5mm}
\section{Results}
\vspace{-1.5mm}

This section describes the results of our study, organized into three subsections. Section~\ref{subsec:baseline_models_and_datasets} (Baseline Models and Datasets) introduces the four baseline models used for comparison with our model. It also details the two real-world datasets on which the various models are evaluated. Section~\ref{subsec:experimental_results} (Experimental Results) describes the performance outcomes of the models on the real-world datasets, highlighting metrics such as accuracy, diversity, and structural measures. Section~\ref{subsec:analysis_and_discussion} (Analysis and Discussion) gives an in-depth analysis of the experimental results. It discusses the trade-offs between model accuracy and dataset diversity, and concludes with the strengths of our model in generating realistic synthetic graphs.

\subsection{Baseline Models and Datasets}
\label{subsec:baseline_models_and_datasets}
\vspace{-0.5mm}

We compare our model to the following four baseline models:
\begin{enumerate}
    \item \textbf{Random:} A random graph model~\cite{random} along with random sampling from the reference feature distribution to generate the flow dataset.
    \item \textbf{Scale-Free (SF):} A scale-free graph model~\cite{scale_free} along with random sampling from the reference feature distribution to generate the flow dataset.
    \item \textbf{Variational Autoencoder (VAE):} A variational autoencoder model that compresses and expands the reference network matrix with some neural loss. This tabular approach is often used with GANs~\cite{gan_netflow2,vae,gan_netflow}.
    \item \textbf{NVIDIA:} A recent model for large scale synthetic graph dataset generation (from a team at NVIDIA)~\cite{nvidia}. Our model is inspired by this approach.
\end{enumerate}
We evaluate these models on netflow datasets from two real-world networks:
\begin{enumerate}
    \item \textbf{MITRE's 5G Core Lab (THOR) network}: The THOR dataset is a 5G core emulation test bed~\cite{Ribeiro2023Enabling5G} that contains multiple clusters and agents to simulate real 5G connections. We sample a 2-hour window of the test bed for our experiments to work with a smaller, more interpretable dataset. This gives us a graph with approximately a thousand flows and two hundred IP addresses.
    \item \textbf{MITRE's Fort Meade Experiment (FMX) environment}: The FMX dataset is observed from a ``living laboratory'' exercise environment within MITRE’s corporate network~\cite{Strom2020MITREATTACK}. The environment is generally used to test adversarial activity on a real-world network. For this work, we extracted only regular activity for the reference network flows. This dataset not only reflects real-world properties but also spans years with many nodes and edges, giving an applicable dataset to test our models ability to scale to larger problems accurately.
\end{enumerate}
In our experiments, we run each model on the THOR and FMX datasets, 20~times for each combination of model and dataset.
For the random and scale-free generators, we used parameters extracted from the reference graph (e.g., number of nodes and edges, and maximum timestamp). For the variational autoencoder, we used a latent space dimension of 3 and trained the model with a learning rate of 0.0001, batch size of 64, and 100 epochs. For the generator called "NVIDIA", we used the parameters described in~\cite{nvidia}.

We use one week of activity from the FMX dataset, giving a graph with approximately 2 million flows and thousands of IP addresses. Thus, the THOR dataset reflects a small snapshot of a large network while the FMX dataset gives a broader picture of all network activity. We cannot disclose the datasets, as they reflect real-world computer network activity, which raises security and privacy concerns. However, the tool described in this paper may enable others to access similar synthetic datasets.

\subsection{Experimental Results}
\label{subsec:experimental_results}
\vspace{-0.5mm}

Table~\ref{tab:thor_results} shows the results from our experiments on the THOR dataset. The Scale-Free (SF) model has the lowest accuracy error (\(\mathcal{A}\)) at 3164, i.e., closer alignment with the reference network. However, it has moderate diversity (\(\mathcal{D}\)) with a value of 31 and a smaller distribution radius (\(\mathcal{R}\)) of 878, suggesting less variability in the generated datasets. Compared to our model, the VAE model exhibits poorer performance in each measure, with higher accuracy error, less diversity, and a smaller radius. The NVIDIA model yields the largest radius \(\mathcal{R}\) at 5823, although it has the poorest diversity, and its accuracy is only moderate. Our model shows a balanced performance, with an accuracy of 4953, diversity of 44, and a radius of 4239, indicating a well-rounded capability in generating diverse yet accurate datasets.


\begin{table}[H]
    \centering
    \caption{Results on the THOR dataset comparing models on accuracy error (\(\mathcal{A}\)), diversity (\(\mathcal{D}\)), and radius (\(\mathcal{R}\)).}
    \label{tab:thor_results}
    \begin{tabular}{ |c||c|c|c|  }
    \hline
    Model & $\mathcal{A} \downarrow$ & $\mathcal{D} \downarrow$ & $\mathcal{R} \uparrow$ \\
    \hline
    Random          & 4298 & \textbf{28} & 2846 \\
    SF              & \textbf{3164} & 31 & 878 \\
    VAE             & 6668 & 45 & 813 \\
    NVIDIA          & 5810 & 82 & \textbf{5823} \\
    \textbf{Ours}   & 4953 & 44 & 4239 \\
    \hline
    \end{tabular}
\end{table}

Table~\ref{tab:fmx_results} shows the results on the FMX dataset. Here, our model excels in both accuracy and diversity, with the lowest \(\mathcal{A}\) value of 827 and the lowest \(\mathcal{D}\) value of 11. The Random model exhibits a high radius (\(\mathcal{R}\)) of 5462986, but this is accompanied by poor accuracy and diversity, reflecting its pure randomness. The Scale Free model achieves a higher radius of 20286 and higher diversity of 273, but has quite poor accuracy (11977). Similarly, the NVIDIA model, while achieving a higher radius of 78396 and higher diversity of 16655, has much poorer accuracy (39881). The VAE model shows overfitting with \(\mathcal{R}\) value approaching zero, indicating a failure to generalize beyond the training data.


\begin{table}[H]
\centering
\caption{Results on the FMX dataset comparing models on accuracy error (\(\mathcal{A}\)), diversity (\(\mathcal{D}\)), and radius (\(\mathcal{R}\)).}
\label{tab:fmx_results}
\begin{tabular}{ |c||c|c|c|  }
\hline
Model & $\mathcal{A} \downarrow$ & $\mathcal{D} \downarrow$ & $\mathcal{R} \uparrow$ \\
\hline
Random          & 3110067 & 1798 & \textbf{5462986} \\
SF              & 11977 & 273 & 20286 \\
VAE             & \textit{595} & \textit{0} & \textit{0} \\
NVIDIA          & 39881 & 16655 & 78396 \\
\textbf{Ours}   & \textbf{827} & \textbf{11} & 453 \\
\hline
\end{tabular}
\end{table}

Table~\ref{tab:thor_struct_results} provides structural measures for the THOR dataset. The table compares models based on degree similarity, average path length, diameter, number of distinct edges, density, and clustering coefficient. Our model demonstrates strong performance in degree similarity (0.898) and clustering coefficient (0.115), closely aligning with the reference network. It also gives the highest average path length (4.01) and diameter (5.32), indicating a more complex network structure. The NVIDIA model achieves the highest degree similarity (0.937) and distinct edges (915), but our model maintains a better balance across all metrics.

\begin{table}[H]
\centering
\caption{Structural measures for the THOR dataset.}
\label{tab:thor_struct_results}
\begin{tabular}{ |c||c|c|c|c|c|c|  }
\hline
Model& Degree & Avg. & Diameter & Edges & Density & Clustering \\
& Sim. & Path &&&& \\
\hline
Random                  & 0.234 & \textbf{3.14} & 3.76 & 1253 & 0.0285 & 0.056  \\
SF                      & 0.650 & 2.69 & 3.66 & 346 & 0.0079 & 0.174  \\
VAE                     & 0.807 & 0.8 & 0.88 & 131 & 0.1321 & 0.0  \\
NVIDIA                  & \textbf{0.937} & 2.64 & 3.47 & \textbf{915} & 0.0364 & 0.292  \\
\textbf{Ours}           & 0.898 & 4.01 & \textbf{5.32} & 692 & \textbf{0.0144} & \textbf{0.115}  \\
\hline
\textit{Reference}      &  1    & 3.3  & 4.56  & 1017 & 0.0232 & 0.140  \\
\hline
\end{tabular}
\end{table}

Table~\ref{tab:fmx_struct_results} details the structural measures for the FMX dataset. Our model achieves the highest degree similarity (0.993), closely matching the reference network, and maintains a reasonable average path length (4.43) and diameter (5.92). The Scale-Free model also performs well, particularly in degree similarity (0.992) and average path length (3.33), but our model shows superior adaptability across different network structures.

\begin{table}[H]
\centering
\caption{Structural measures for the FMX dataset.}
\label{tab:fmx_struct_results}
\begin{tabular}{ |c||c|c|c|c|c|c|  }
\hline
Model& Degree & Avg. & Diameter & Edges & Density & Clustering \\
& Sim. & Path &&&& \\
\hline
Random                  & 0.040 & 1.77 & 1.87 & 1217934 & 0.227 & 0.402  \\
Scale-Free              & 0.992 & \textbf{3.33} & \textbf{4.36} & 4270 & \textbf{0.0008} & 0.142  \\
VAE                     & 0.681 & 0.47 & 0.79 & 16   & 0.0523 & \textbf{0.0}  \\
NVIDIA                  & 0.860 & 4.66 & 6.24 & 518 & 0.0035 & 0.026  \\
\textbf{Ours}           & \textbf{0.993} & 4.43 & 5.92 & \textbf{3619} & 0.001 & 0.219  \\
\hline
\textit{Reference}      &  1    & 3.13  & 3.79  & 3619 & 0.0007 & 0.0  \\
\hline
\end{tabular}
\end{table}

\begin{figure}[!htbp]
    \centering
    \subfloat[\centering VAE model]{{\includegraphics[width=10.7cm]{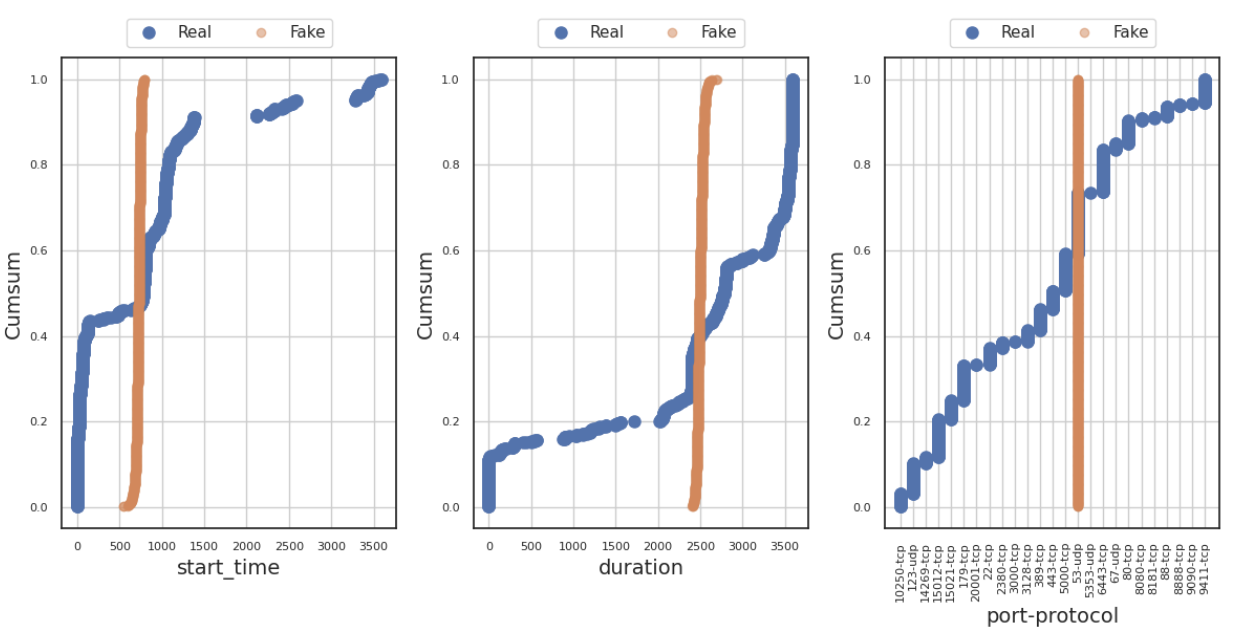}}}%
    \qquad
    \subfloat[\centering NVIDIA model]{{\includegraphics[width=10.7cm]{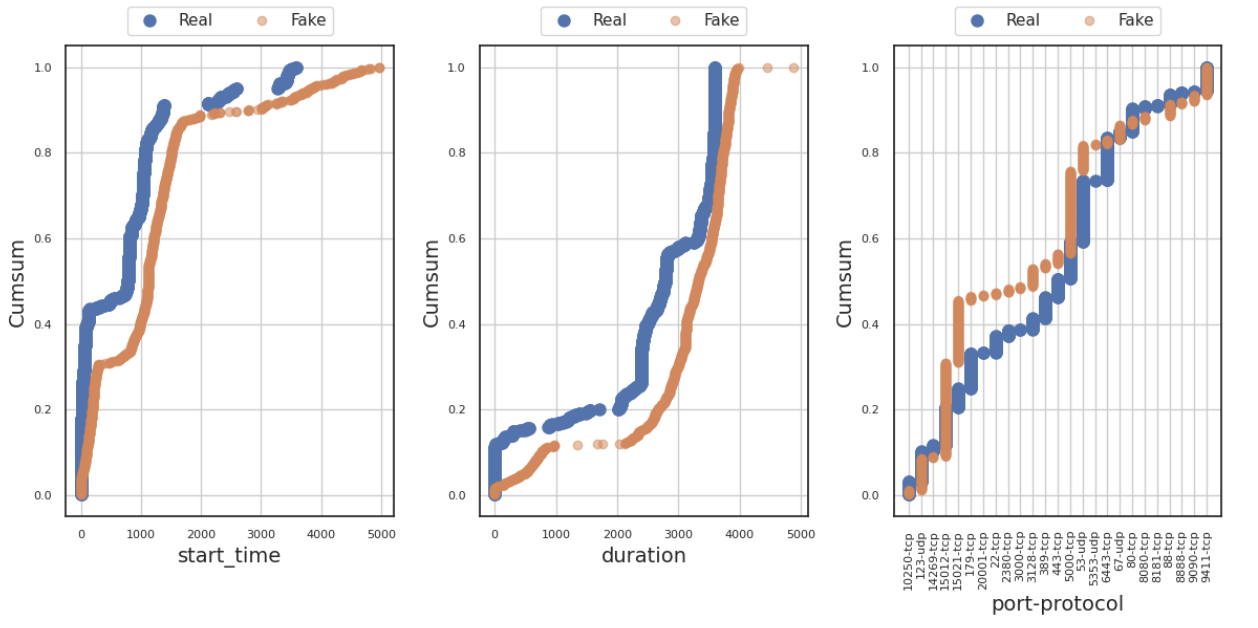}}}%
    \qquad
    \subfloat[\centering Our model]{{\includegraphics[width=10.7cm]{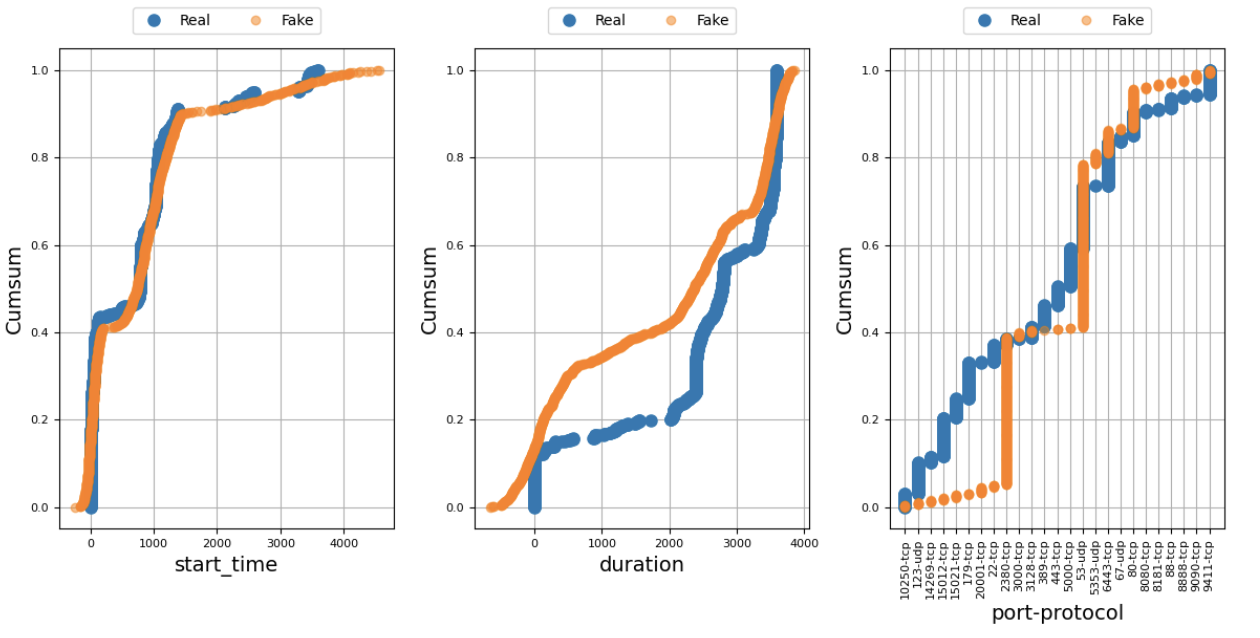}}}%
    \caption{CDF plots of netflow feature distributions for the THOR dataset. These compare the alignment of start-time, duration, and port-protocol features with the reference data.}%
    \label{fig:thor_feat_results}%
\end{figure}

\begin{figure}[!htbp]
    \centering
    \subfloat[\centering VAE model]{{\includegraphics[width=10.7cm]{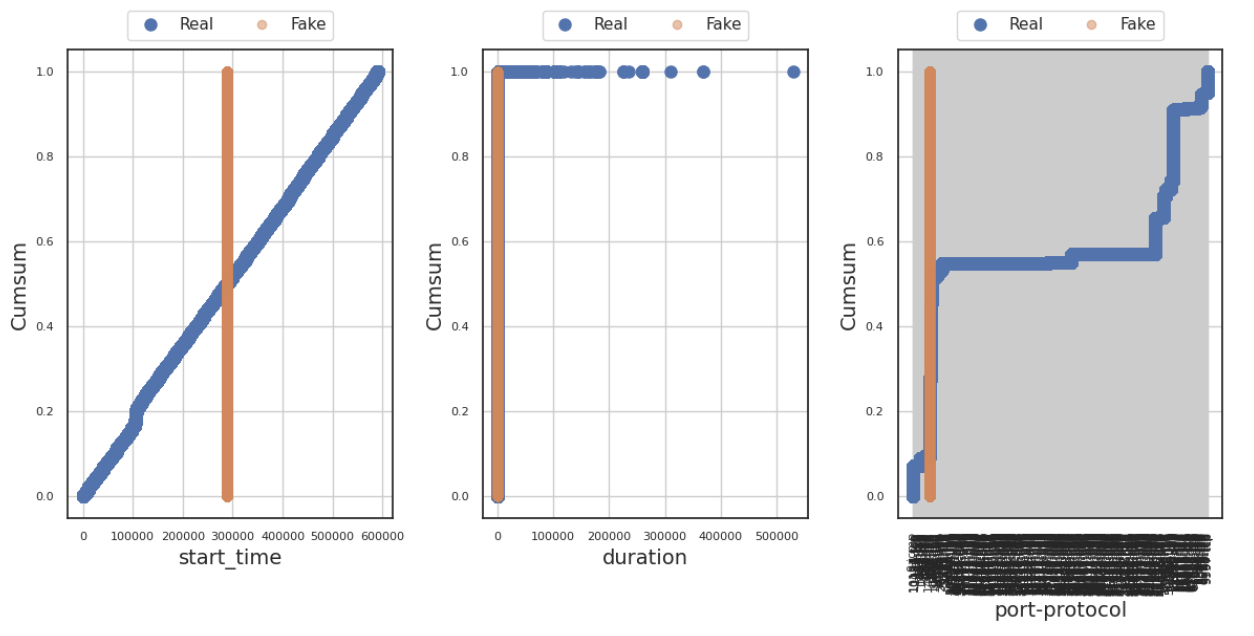}}}%
    \qquad
    \subfloat[\centering NVIDIA model]{{\includegraphics[width=10.7cm]{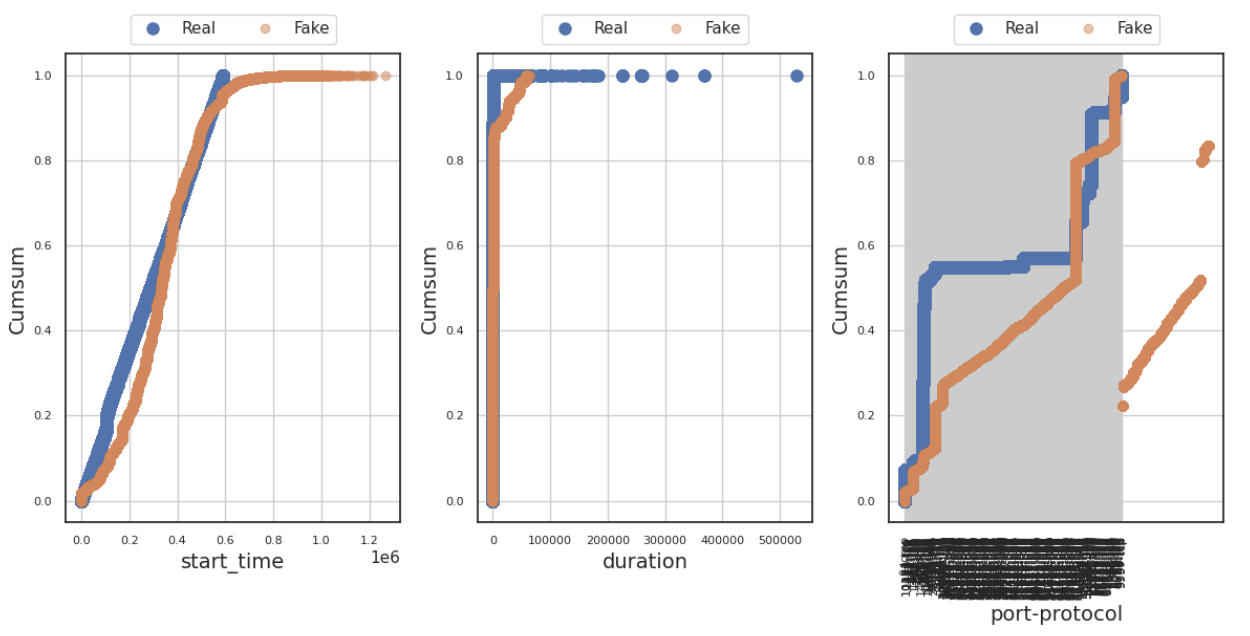}}}%
    \qquad
    \subfloat[\centering Our model]{{\includegraphics[width=10.7cm]{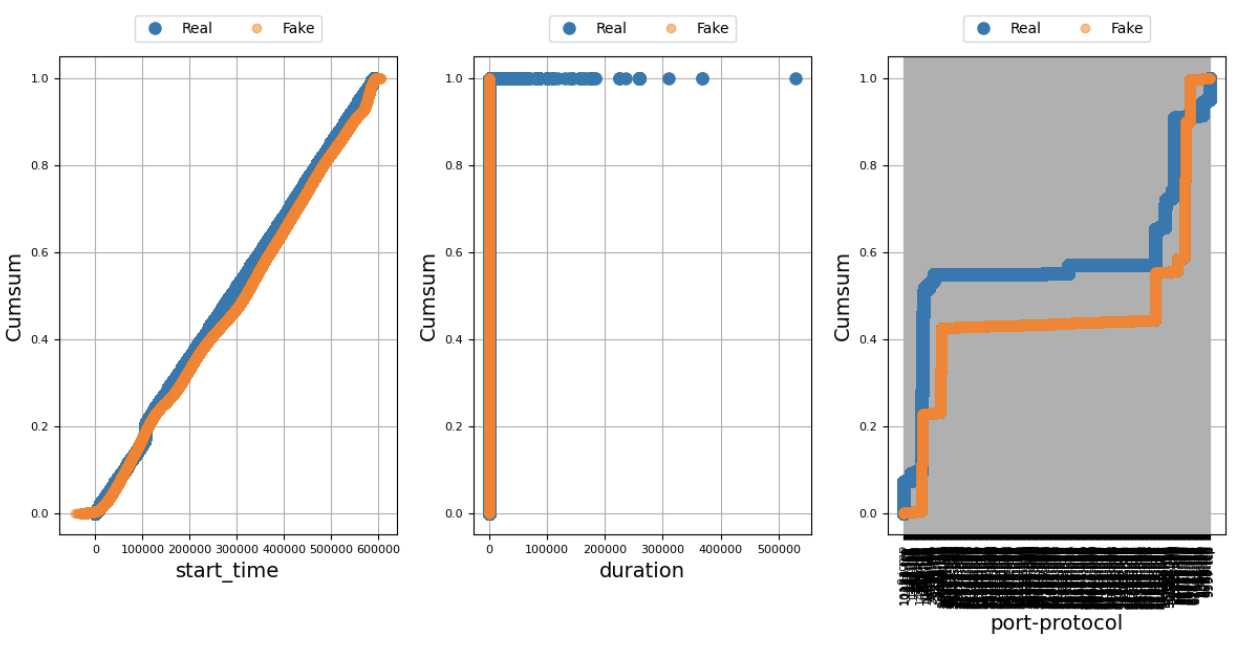}}}%
    \caption{CDF plots of netflow feature distributions for the FMX dataset. These compare the alignment of start-time, duration, and port-protocol features with the reference data.}%
    \label{fig:fmx_feat_results}%
\end{figure}

Figures \ref{fig:thor_feat_results} and \ref{fig:fmx_feat_results} present CDF plots for the THOR and FMX datasets, respectively. These plots illustrate the distributions of netflow-based features—start-time, duration, and port-protocol—which are non-topological aspects of the data, i.e., distinct from nodes (IP addresses) and edges (network flows). The CDF plots focus on the models with non-random feature sampling: VAE, NVIDIA, and ours. These models are designed to learn and replicate the distribution of features from the reference data. In contrast, the random and scale-free models rely on random sampling from the reference feature distribution without learning underlying patterns.

In Figure~\ref{fig:thor_feat_results} (for THOR), both our model and the NVIDIA model achieve similar alignment with the real data across the netflow features. For start time, both models closely match the reference data, though the NVIDIA model shows a slight deviation. In the duration feature, the NVIDIA model underestimates flow durations, while our model slightly overestimates them. Despite these differences, both models maintain comparable overall alignment. For port-protocol, our model has similar performance as the NVIDIA one in replicating the distribution. In contrast, the VAE model significantly under performs, showing large deviations across all features.

In Figure~\ref{fig:fmx_feat_results} (for FMX), our model demonstrates a nearly identical match to the reference data for both the start-time and duration features, indicating a high level of accuracy in capturing these aspects of netflow data. As the figure shows, longer duration flows are infrequent in the reference data. The NVIDIA model, while generally aligned, shows slight deviations for these features, suggesting minor discrepancies. The VAE model strongly deviates for the start-time feature but achieves a nearly identical match for duration, indicating inconsistent performance across features. For the port-protocol feature, our model's curve closely matches the shape of the reference, although it is somewhat below the reference, indicating a slight underestimation. In contrast, the port-protocol features for the NVIDIA and VAE models show almost no similarity to the reference, highlighting a major gap in their ability to replicate this aspect of the data.

\begin{figure}[!htbp]
    \centering
    \includegraphics[width=11cm]{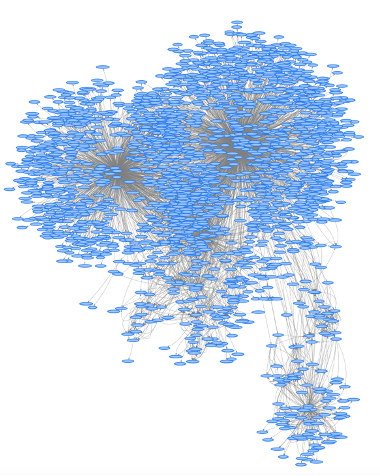}
    \caption{Visualization of the FMX reference netflow graph. This graph represents the real-world network structure used as a benchmark for evaluating the accuracy and diversity of synthetic graph generation models. Nodes and edges illustrate the connectivity and flow patterns within the network.}
    \label{fig:ref_network}
\end{figure}

\begin{figure}[!htbp]
    \centering
    \includegraphics[width=12.76cm]{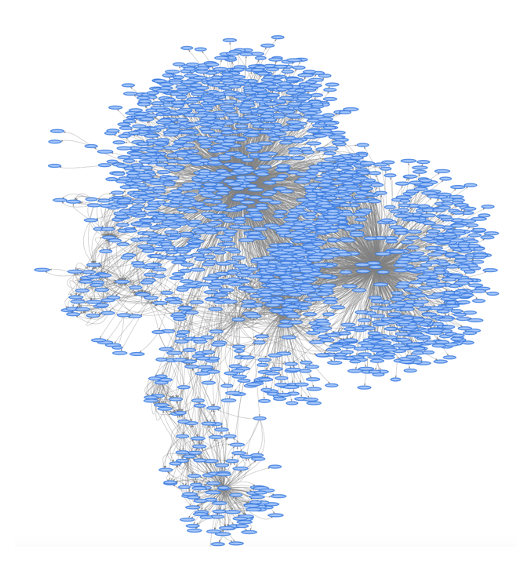}
    \caption{Visualization of a synthetic FMX netflow graph generated by our model. This graph demonstrates the model's ability to learn the structure and flow patterns of the reference network, highlighting its effectiveness in producing realistic synthetic datasets.}
    \label{fig:syn_network}
\end{figure}

Figure~\ref{fig:ref_network} is a visualization of the FMX reference netflow graph. Figure~\ref{fig:syn_network} is a visualization of a synthetic netflow graph generated through our approach. These visualizations are generated with the CyGraph tool~\cite{cygraph}.

\subsection{Analysis and Discussion}
\label{subsec:analysis_and_discussion}
\vspace{-0.5mm}

The VAE model performs decently on the smaller THOR network, but overfits for a larger one (i.e., for FMX). This is supported by Table~\ref{tab:fmx_struct_results}, where we see that the model only finds 16 distinct edges to be useful and disregards the rest. This also occurs in the feature distributions in Figures~\ref{fig:thor_feat_results} and~\ref{fig:fmx_feat_results}, where only a certain type of netflow is generated. This confirms the claim that tabular methods fail when trying to generate graph-like data.

Regarding graph generators, over both domains, ours has the most consistent structural and feature results with the reference network. While the VAE model overfits for the FMX dataset, our model has the best $\mathcal{A}$ and $\mathcal{D}$. However, this leads to a smaller radius because the model is more accurate and fine-tuned.

This is evidence of a trade-off between accurate models and diverse datasets in the current work for scalable graph generation. It shows that none of these models are near the optimal approach, as an ideal model would be able to increase the radius without foregoing accuracy. Therefore, although the NVIDIA model has a better $\mathcal{R}$, we theorize that this is solely because both models are subject to this trade-off; we outperform it on almost all the other measures.

Additionally, the scale-free generator works decently on the FMX dataset, but poorly on the THOR dataset. We claim that it performs poorly on THOR, even though the $\mathcal{A}$ value is the smallest, because it generates very few edges compared to other models. Since the adjacency matrices are sparse, by having fewer edges there is a higher chance of matching the reference graph. Thus, it is insufficient to look merely at the primary metrics, which is why we include the secondary measures. The difference in the scale-free generator's performance between domains is likely because the FMX dataset follows a scale-free pattern as shown in Figure~\ref{fig:ref_network} and Figure~\ref{fig:syn_network}. However, this model is not generalizable to domains without this property, which occurs in domains like THOR\@.

Overall, our model is the most suitable for real-world synthetic graph generation, as it can scale to larger graphs without losing information, can move between domains that have different underlying assumptions, and is more accurate than previous work that satisfies these criteria.

We also introduce unbiased joint structure and dynamics evaluation metrics for both dataset accuracy and diversity. These metrics are still consistent with the current standard for graph evaluation as they both reach the same conclusions, but also provide more comprehensive evaluation criteria. This highlights the inability of any current large graph dataset synthesizer to have both optimal accuracy and diversity, which suggests new approaches should be explored. However, they are not standalone metrics as factors such as the sparsity of the adjacency matrices could be exploited without having secondary measures for support.

\vspace{-1.5mm}
\section{Conclusion}
\vspace{-1.5mm}

The objective of our work is to develop a tool that can create synthetic, scalable, and diverse cyber network activity given a single reference network. We applied our approach for generating network flows by structuring it as a graph generation problem, which outperforms tabular approaches when moving to larger networks. We took the framework in~\cite{nvidia} and extended it with a more accurate structure generator and aligner, which allows it to work better with dynamic multigraphs. While we utilize more complex models, they still have similar time complexity, allowing for scalability.

We also find that only a small subset of models are able to generate graphs in a scalable way with features without prior assumptions; we outperform those that do in terms of accuracy. Further, we introduce new evaluation metrics that combine structure and features for both accuracy and diversity of synthetic datasets. Using these, we determine that there is a trade-off between accuracy and diversity for the models (including ours). This suggests that there is room for model improvement, motivating further research.

Overall, our work has improved and extended synthetic graph dataset generation, while providing a new tool for gaining access to data without about privacy concerns. Our model is not restricted to netflows, and can work with other graph-based events such as authentication events. Future work includes finding ways to increase diversity without losing accuracy, and incorporating adversarial activity in dataset generation.

\section*{Acknowledgments}

This research was funded by MITRE's Independent Research and Development Program. 


{
\small

\bibliographystyle{plain}
\bibliography{citations}

\begin{thebibliography}{10}

\bibitem{scale_free}
R\'eka Albert and Albert-L\'aszl\'o Barab\'asi.
\newblock Topology of evolving networks: Local events and universality.
\newblock {\em Phys. Rev. Lett.}, 85:5234--5237, Dec 2000.

\bibitem{rmat}
Deepayan Chakrabarti, Yiping Zhan, and Christos Faloutsos.
\newblock {R-MAT}: A recursive model for graph mining.
\newblock In {\em SIAM International Conference on Data Mining}, 2004.

\bibitem{Chen2016XGBoost}
Tianqi Chen and Carlos Guestrin.
\newblock {XGBoost}: A scalable tree boosting system.
\newblock In {\em Proceedings of the 22nd ACM SIGKDD International Conference
  on Knowledge Discovery and Data Mining}, KDD '16, page 785–794. Association
  for Computing Machinery, 2016.

\bibitem{nvidia}
Sajad Darabi, Piotr Bigaj, Dawid Majchrowski, Artur Kasymov, Pawel Morkisz, and
  Alex Fit-Florea.
\newblock A framework for large-scale synthetic graph dataset generation.
\newblock {\em IEEE Transactions on Neural Networks and Learning Systems},
  pages 1--11, 2025.

\bibitem{random}
Paul Erd\"{o}s and Alfr\'{e}d R\'{e}nyi.
\newblock On random graphs {I}.
\newblock {\em Publicationes Mathematicae Debrecen}, 6:290, 1959.

\bibitem{gnn3}
Faezeh Faez, Yassaman Ommi, Mahdieh~Soleymani Baghshah, and Hamid~R. Rabiee.
\newblock Deep graph generators: A survey, 2020.

\bibitem{gnn}
Xiaojie Guo and Liang Zhao.
\newblock A systematic survey on deep generative models for graph generation,
  2022.

\bibitem{kron}
Jure Leskovec, Deepayan Chakrabarti, Jon Kleinberg, Christos Faloutsos, and
  Zoubin Ghahramani.
\newblock Kronecker graphs: An approach to modeling networks, 2009.

\bibitem{gnn2}
Yujia Li, Oriol Vinyals, Chris Dyer, Razvan Pascanu, and Peter Battaglia.
\newblock Learning deep generative models of graphs, 2018.

\bibitem{diffusion}
Chengyi Liu, Wenqi Fan, Yunqing Liu, Jiatong Li, Hang Li, Hui Liu, Jiliang
  Tang, and Qing Li.
\newblock Generative diffusion models on graphs: Methods and applications,
  2023.

\bibitem{cygraph}
Steven Noel, Eric Harley, Kam~Him Tam, Michael Limiero, and Matt Share.
\newblock {CyGraph}: Graph-based analytics and visualization for cybersecurity.
\newblock In {\em Cognitive Computing: Theory and Applications}, volume~35 of
  {\em Handbook of Statistics}, pages 117--167. Elsevier, 2016.

\bibitem{Ribeiro2023Enabling5G}
Leila~Z. Ribeiro, Thierry Klein, Luiz~A. DaSilva, Attila Takacs, Prasanth
  Ananth, Salvatore D’Oro, Dipesh Modi, Rick Niles, and Izabela Gheorghisor.
\newblock Enabling {5G} innovation leadership through use case-driven
  collaboration.
\newblock Technical report, Open Generation Consortium, 2023.

\bibitem{gan_netflow2}
Markus Ring, Daniel Schlör, Dieter Landes, and Andreas Hotho.
\newblock Flow-based network traffic generation using generative adversarial
  networks.
\newblock {\em Computers \& Security}, 82:156--172, May 2019.

\bibitem{vae}
Martin Simonovsky and Nikos Komodakis.
\newblock {GraphVAE}: Towards generation of small graphs using variational
  autoencoders, 2018.

\bibitem{Strom2020MITREATTACK}
Blake Strom, Andy Applebaum, Doug Miller, Kathryn Nickels, Adam Pennington, and
  Cody Thomas.
\newblock {MITRE ATT\&CK\textsuperscript{\textregistered}}: Design and
  philosophy.
\newblock Technical Report MP180360R1, MITRE Corporation, 2020.

\bibitem{lanl}
Melissa J.~M. Turcotte, Alexander~D. Kent, and Curtis Hash.
\newblock Unified host and network data set, 2017.

\bibitem{small_world}
Duncan~J. Watts and Steven~H. Strogatz.
\newblock Collective dynamics of ‘small-world’ networks.
\newblock {\em Nature}, 393(6684):440--442, 1998.

\bibitem{ctgan}
Lei Xu, Maria Skoularidou, Alfredo Cuesta-Infante, and Kalyan Veeramachaneni.
\newblock Modeling tabular data using conditional {GAN}, 2019.

\bibitem{gan_netflow}
Yucheng Yin, Zinan Lin, Minhao Jin, Giulia Fanti, and Vyas Sekar.
\newblock Practical {GAN}-based synthetic {IP} header trace generation using
  {NetShare}.
\newblock In {\em Proceedings of the ACM SIGCOMM 2022 Conference}, pages
  458--472, 2022.

\bibitem{dynamic}
Wenbin Zhang, Liming Zhang, Dieter Pfoser, and Liang Zhao.
\newblock Disentangled dynamic graph deep generation, 2021.

\end{thebibliography}
}

\end{document}